\documentclass[journal]{IEEEtran}

\usepackage{color}
\usepackage{epsf}
\usepackage{times}
\usepackage{epsfig}
\usepackage{epstopdf}
\usepackage{graphicx}
\usepackage{algorithm}
\usepackage{algorithmic}
\usepackage{amsmath}
\usepackage{amssymb}
\usepackage{amsxtra}
\usepackage{multirow}
\usepackage{mathtools}
\usepackage{subcaption}
\usepackage{multirow}
\usepackage{comment}
\usepackage[normalem]{ulem}
\usepackage{tabu}
\usepackage[dvipsnames]{xcolor}
\usepackage{multicol}
\usepackage{balance}
\usepackage{cite}
\usepackage{caption}
\begin{document}

\title{\vspace{-0.35cm}Service Discovery in Social Internet of Things using Graph Neural Networks\vspace{0cm}}

\author{\IEEEauthorblockN{Aymen Hamrouni, \textit{Student Member, IEEE}, Hakim Ghazzai, \textit{Senior Member, IEEE} \\ and Yehia Massoud, \textit{Fellow, IEEE}}
{\thanks {\vspace{-0.3cm}\hrule \vspace{0.1cm}
Aymen Hamrouni, Hakim Ghazzai, and Yehia Massoud are with King Abdullah University of Science and Technology (KAUST), Thuwal, Makkah, KSA. (E\textendash mails: \{aymen.hamrouni, hakim.ghazzai, yehia.massoud\}@kaust.edu.sa).\newline
This paper is accepted for publication in 65th IEEE International Midwest Symposium on Circuits and Systems (MWSCAS'22), Virtual Conference, Aug. 2022. \newline \textcopyright~2022 IEEE. Personal use of this material is permitted. Permission from IEEE must be obtained for all other uses, in any current or future media, including reprinting/republishing this material for advertising or promotional purposes, creating new collective works, for resale or redistribution to servers or lists, or reuse of any copyrighted component of this work in other works.
}}
\vspace{-0.5cm}}

\maketitle

\begin{abstract}
\boldmath{Internet-of-Things (IoT) networks intelligently connect thousands of physical entities to provide various services for the community. It is witnessing an exponential expansion, which is complicating the process of discovering IoT devices existing in the network and requesting corresponding services from them. As the highly dynamic nature of the IoT environment hinders the use of traditional solutions of service discovery, we aim, in this paper, to address this issue by proposing a scalable resource allocation neural model adequate for heterogeneous large-scale IoT networks. We devise a Graph Neural Network (GNN) approach that utilizes the social relationships formed between the devices in the IoT network to reduce the search space of any entity lookup and acquire a service from another device in the network. This proposed approach surpasses standardization issues and embeds the structure and characteristics of the social IoT graph, by the means of GNNs, for eventual clustering analysis process. Simulation results applied on a real-world dataset illustrate the performance of this solution and its significant efficiency to operate on large-scale IoT networks.}
\end{abstract}


\section{Introduction}
\label{Sec1a}
The Internet-of-Things (IoT) is the network of various physical objects or “things” encompassed with different software and various types of sensors  (e.g., weather sensors, smart traffic lights) along with other capabilities (e.g., actuators, computational resources). Heterogeneous devices in the IoT network offer a wide range of services with different functionalities and performances. They interconnect, share data, and exchange these services among each other to achieve more value and enable different distributed paradigms. Over the last decade, the number of these devices has been astronomically increasing due to the rapid proliferation and the abundant availability of wearables, smart sensors, and smartphones along with the advent and the development of different paradigms and technologies associated with IoT such as smart traffic management, healthcare systems, and mobile crowdsourcing applications~\cite{9474925,9140078}. 

In the IoT network, an entity may need to perform certain tasks that go beyond its capabilities (e.g., seek additional computational resource during high server demand). To find a suitable candidate capable of successfully achieving such tasks, the IoT devices resort to service discovery processes where they search for appropriate services and their providers in the IoT network while taking into account their requests’ context and the needed Quality of Service (QoS)~\cite{9474925}. However, it remains challenging to locate desirable services in such ubiquitous and vast networks. Moreover, the IoT services are deployed in resource-constrained devices that are characterized by their mobility nature and limited capabilities in terms of battery, computing, communication, and storage. Hitherto, the traditional service discovery solutions are not suitable for effective exploitation of IoT services as affirmed by several studies such as in \cite{8913969}. 
The key challenge is the lack of IoT standardization in terms of architectures and new technologies adopted to meet the IoT needs. Also, the service exchanges are operated in a highly dynamic environment with a massive number of devices. Therefore, an effective mechanism of service discovery is needed to meet the requirements of both IoT environment and IoT services.

In this context, combining the Graph Neural Networks paradigm (GNNs) with the Social IoT (SIoT) concept could be a promising solution to provide an efficient service discovery~\cite{9446513}. The SIoT refers to the social relationships established between IoT devices. It mainly combines social networks with IoT, given that objects, similar to humans, can be considered intelligent and cooperative, and consequently they can create their own social network graph to interact with each other and achieve their common goals. On the other hand, GNNs are neural models that operate on such graphs and explore their structure and the dependence between the nodes. As the SIoT relationships between devices can provide more information about the nature of connections between the entities such as compatibility and trustworthiness, Utilizing GNN to analyze and utilize these relationships can provide an automated solution for such service discovery problems~\cite{9142908,9345852}. GNNs can analyze the semantic of the SIoT graph in a time-efficient way. 

In the literature, there are only few research work that studied service discovery while relying on SIoT. In~\cite{9181080,doi:10.1080/23335777.2019.1678198}, some works designed an automated service discovery process to allow mobile crowdsourcing task requesters select a small set of devices out of a large-scale Internet-of-things (IoT) network to execute their tasks. Other studies
presented a simple algorithm to discover resources/services among SIoT communities. They discussed community detection within established SIoT network and studied intracommunity and intercommunity service search algorithms for efficient service discovery among the SIoT communities. 

In this paper, we propose the use of GNNs to enable efficient and rapid service discovery in the dynamic large-scale Social IoT networks while capturing not only the  social relations in the network, but also the devices' features (e.g., mobility, computational capabilities, etc). Afterward, and based on the SIoT relations interconnecting the different devices, we cluster the network into several groups with strong social relations. We show that these approaches while relying on several SIoT relationships, help reduce the service lookup search space and consequently speed up the service discovery process. We perform a comparison between different embedding strategies: using edges only, features only, and a combination of both.

\section{Service Discovery in SIoT}\label{sec3}



Since many heterogeneous objects offer different services with different exchange protocols, communication channels, and in a widely distributed manner, it is both challenging and very complex to locate desirable services in a short time. 
In this paper, we propose to use SIoT, which is a concept that mimics the structure of the human social network and establish different type of relationships between objects. Defining such relationships between objects can reflect much more information about the topology and connections between the IoT devices and eventually assist in the search procedure of the relevant services from one object to another in the IoT network. There are a variety of type of relationships that can be defined between the objects such as Parental–child relation (PCR), Co-Location/Co-Work based Relation (CLOR), Social Object Relation (SOR), and Object Ownership Relation (OOR). We have combined the SOR and OOR relations and proposed a new relationship called Social Friendship and Ownership Relation (SFOR). In a nutshell, the SFOR relation is established by considering the social relationships of the owners of the IoT devices. For example, two devices having the same owner are assumed to have an SFOR relationship.

Let $\mathcal G(\mathcal W,\mathcal E)$ be the graph modeling the SIoT network formed by the devices in the IoT network. Every vertex of $\mathcal G$ corresponds to an IoT device $w \in \mathcal W$ while the set of edges $ \mathcal E$ represents the SIoT relationships between the workers. Let $\mathbf{F}_{w}$ be the feature vector of device $w$ in the graph. $\mathbf{F}_{w}$is numerical vector encompassing the characteristics and capabilities of node $w$ such as mobility, battery, etc. In service discovery, the service requester $w_r$, having specifications $\mathbf{F}_{w_r}$ provides the required characteristics $\mathbf{F}_{w_p}$in the search of a suitable IoT device $w_p$. Such a problem can be modeled as a combinatorial multi-objective optimization where the optimal service provider is the IoT device that is closest to the service provider in the SIoT network and offering the required characteristics. As this problem is NP-hard~\cite{9181080}, we propose a smart service lookup framework that maps the request of the service requester and identify potential candidates.

\begin{figure}[t]
\centering
        \includegraphics[width=9.25cm]{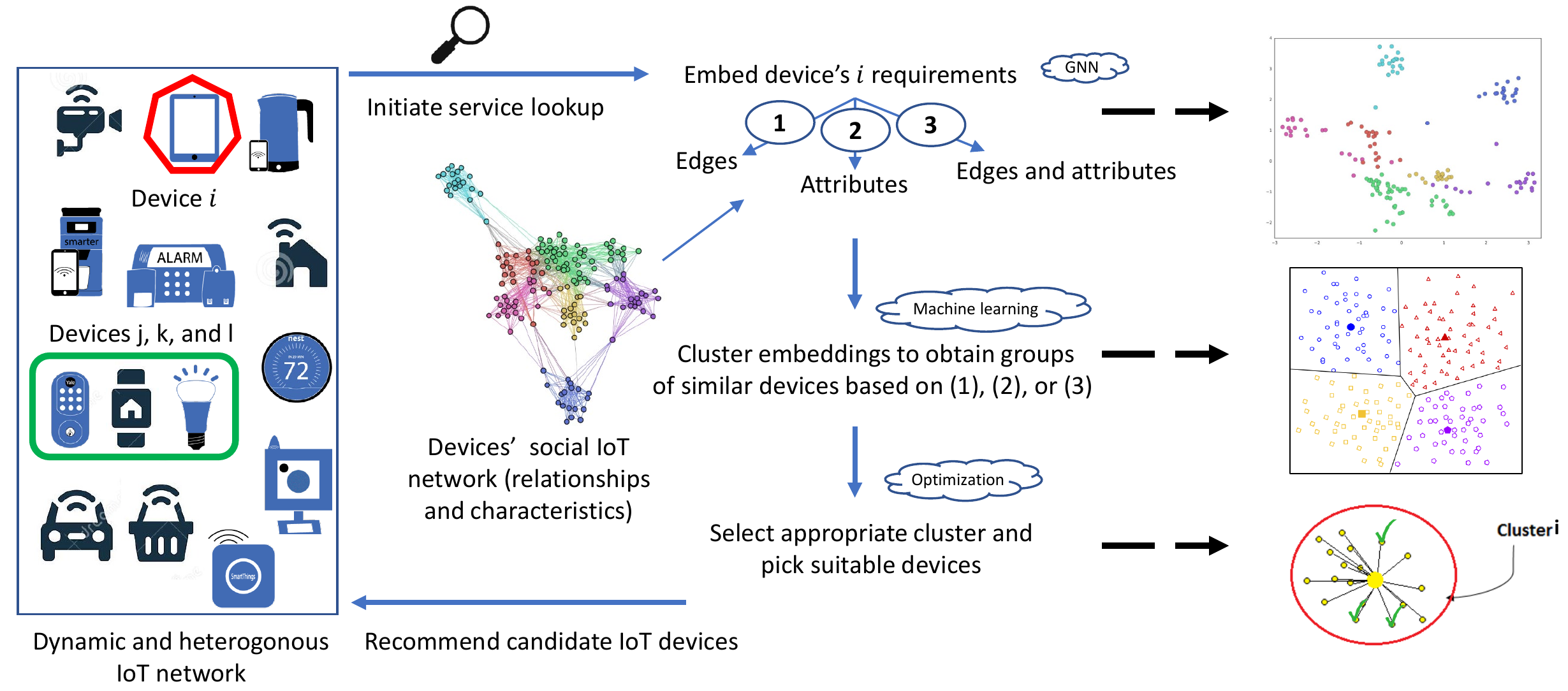}
    \caption{Architecture of the proposed service lookup framework to map the request of an IoT device in the network and identify potential candidates.}
    \label{arch}
    \vspace{-0.2cm}
\end{figure}

As Fig~\ref{arch} shows, initially, the service requester $w_r$ initiates a service lookup procedure while providing  $\mathbf{F}_{w_p}$. These requirements are embedded by the means of a GNN technique and eventually, a grouping technique is initiated to group similar nodes' characteristics together. Depending on the embedding approach and how the embedding is performed, the clustering phase yields candidates devices as service providers.

In the following sub-sections, we introduce the embedding, the clustering, and the service lookup phases.

\subsection{Embedding}
In order to achieve successful service discovery with minimal search time and lower complexity. we harness the power of GNNs to embed the total SIoT network to lower dimensional space. By relying on GNN, we model the graph structure and the attributes of the IoT devices (e.g., mobility, battery information, model, etc.) in one dimensional space. The output of embedding is a vector space containing information about the IoT devices, their relationships, and their characteristics. Similar nodes in the graph are embedded close to each other.

The proposed solution embeds the SIoT graphs using three different approaches: i) embeds the devices' relationships only, ii) embeds the devices' attributes only, and iii) embeds the devices' relationships and attributes together. In the first approach, the GNN learns only about the structure and the topology of the SIoT graph and the resultant embedding reflects only the relationships of the devices. In the second approach, and since there are no edges, the GNN is reduced to a neural network that learns the features of the devices. In the third approach, the resultant embedding takes into consideration, not only the connections between the devices, but also their features and characteristics. 

The way embeddings are generated from the graph is by defining at first a strategy $R$ that runs flexible and biased random walks starting from each device in the graph to other devices in the graph. This strategy allows nodes to be discovered and enable the GNN to determine which nodes are neighbors with which nodes. While walking from a node to the other, features information are passed and aggregated in the neural network. In all of the three approaches, the overall goal of the GNN is to minimize the likelihood of two similar nodes in the graph. The final loss function which we optimize is defined as follows :
\begin{equation}
\mathcal{L}=\sum_{w \in \mathcal{W}} \sum_{w' \in N_{R}(w)}-\log \mathrm{P}\left(w'/ \mathbf{F}_{w}\right)
\end{equation}
where $N_{R}(w)$ is the multiset of nodes visited on random walks starting from $w'$ to node $w$. The term
$\mathrm{P}\left(w' / \mathbf{F}_{w}\right)$ represents the likelihood of random walk co-occurrences and it is expressed as follows:
$\mathrm{P}\left(w' /\mathbf{F}_{w}\right)=\frac{\exp \left(\mathbf{F}_{w}^{\top} \mathbf{F}_{w'}\right)}{\exp \left(\sum_{\mathbf{n} \in \mathcal{W}} \mathbf{F}_{w}^{\top} \mathbf{F}_{n} \right)}$.

\subsection{Clustering}
After embedding the graph, we obtain, for all of the three different approaches, lower dimensional vectors containing the embedded information about the IoT devices. Since ideally, an embedding captures the semantics of the input by placing similar inputs close together in the embedding space, we propose to leverage this aspect and proceed with a clustering analysis technique that highlights groups of devices that are close to each other. Throughout the literature, there have been a wide variety of possible machine learning approaches for clustering such as Agglomerative Clustering, DBSCAN, $K$-Means, Spectral Clustering, and Mixture of Gaussians. The clustering returns a list of groups of of devices sharing a certain level of similarities, in our context, similar embedding.

\subsection{Service Lookup}
After performing the clustering, we initiate the service lookup phase where we start searching for a suitable IoT device to provide the required service.

\textbf{Service lookup with edge embedding only:} as the resultant clusters contain devices that are interconnected in the SIoT graph, we look for the cluster that contains the service requester. That cluster, indeed, contains the most socially connected devices to the requester, i.e., trustworthy but without any knowledge about their characteristics. Therefore, we explore all of the devices in that cluster and choose the one that is most relevant to the required service $\mathbf{F}_{w}$. In this case, the search space is severely reduced from going through all of available devices in the entire network to only a small set of similar devices having strong relations with the service requester. The device that which has the closest characteristics to the required service in terms of euclidean distance is the best suitable candidate to achieve the service. 

\begin{figure}[t]
\vspace{-0.3cm}
\hspace{-0.5cm}
\subfloat[]{     \includegraphics[width=4.5cm]{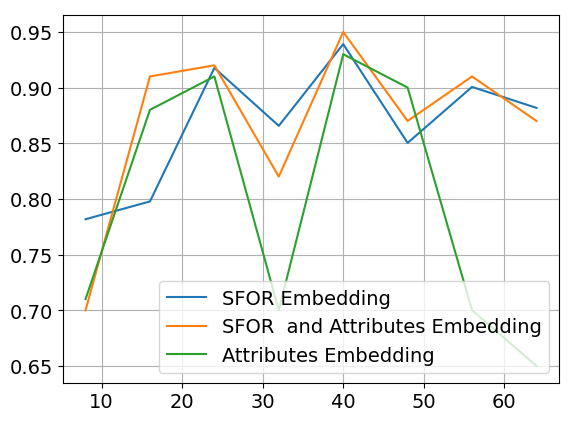}
}
\subfloat[]{  \includegraphics[width=4.5cm]{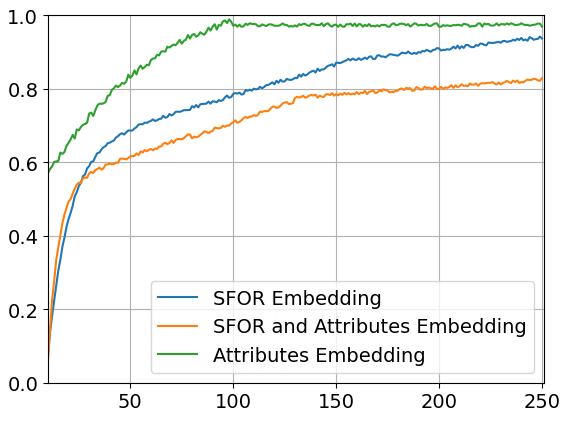}}
    \caption{(a) Accuracy of embedding technique vs. the number of embedding dimension (b) Evolution of the embedding accuracy from epoch to epoch for the SFOR relations embeddings, the attributes embedding, and the SFOR and attributes embedding}
  \label{2other}
  \vspace{-0.4cm}
\end{figure}

\textbf{Service lookup with features embedding only:} the resultant clusters contain the most close devices to each other in terms of characteristics. However, since we are looking for a specific service provider with certain characteristics, we choose to implement a fake device at the original SIoT network that has the desired characteristics. In this way, and after clustering, we look for the cluster that contains this fake node. All the devices in that cluster are potential candidates for the service requester. To find the most suitable one, we search for the device with the closest relationship to the service requester. This can be done using the Dijkistra algorithm where we compute the shortest path for each of the devices in cluster. 

\textbf{Service lookup with full embedding:} like the previous case, we introduce a fake device at the original SIoT network that has the desired characteristics but this time with also the service requester's social relationships. After the clustering process, we look for the cluster that contains this fake node. All the devices in that cluster are potential candidates for the service requester. In other words, they are the closest ones to the desired characteristics with strong relationships with the service requester. The device with the closest characteristics and social relationships to the required service is the best suitable candidate to accomplish the desired service.

\section{Experiments and Evaluation}\label{sec4}

\subsection{Experimental Setup}
To evaluate the performances of the proposed service allocation approach, we use a public SIoT dataset made available by Marche et al. ~\cite{marche2020exploit}. It includes real IoT objects located in Santander, Spain, mixed with artificial data objects such as smartphones, tablets, and personal computer devices. The total number of objects is $16,216$ devices, of which $14,600$ are from private users and $1,616$ from public services. For each device, the database includes several information such as the device's type, its mobility, the brand, and its model. For private IoT devices, additional information about the owner identification is also stored. For all of our simulations, we perform a Monte Carlo simulation with 500 iteration where each time we randomly choose $933$ private static IoT objects. We also focus only on four of the device's attributes, namely their type, brand, mobility, and power supply nature (i.e., runs on battery or not). Moreover, only these attributes are one-hot encoded and later used to characterize the IoT device's features and establish the attribute embedding. In each of the 500 Monte Carlo simulations, we create the SFOR relationships for the extracted $933$ sample IoT devices using the device's owners. All the discussed algorithms are implemented in a Python 3.7 environment and on a 32 socket Intel(R) Xeon (R) E5-2698 v3 @2.30GHz CPU with 72G of RAM.

\begin{table*}[t]
\vspace{-0.5cm}
\centering
\caption{Characteristics of the embedding approaches.}
\label{tablescaract}
\resizebox{15cm}{!}{%
\addtolength{\tabcolsep}{-0pt} \scalebox{0.3}{\begin{tabular}{c|c|c|c|}
\cline{2-4}
                                                                                                                          & Edges Embedding & Attributes Embedding & Edge-Attributes Embedding \\ \hline
\multicolumn{1}{|c|}{Minimum Number of Epochs for 95\%}                                                                   & 250             & 120                  & 730                       \\ \hline
\multicolumn{1}{|c|}{Accuracy (Epoch=500)}                                                                                            & 0.95           & 0.98                 & 0.91                      \\ \hline
\multicolumn{1}{|c|}{Average Number of Candidate Devices}                                                                 & 360             & 18                   & 140                       \\ \hline
\multicolumn{1}{|c|}{\begin{tabular}[c]{@{}c@{}}Similarity to Service requester's \\ SIoT Relations (\%)\end{tabular}}         & 89            & 52                 & 72                      \\ \hline
\multicolumn{1}{|c|}{\begin{tabular}[c]{@{}c@{}}Similarity to Service requester's\\ Required Caracteristics (\%)\end{tabular}} & 71            & 91                 & 88                      \\ \hline
\end{tabular}}}
\vspace*{-0.3cm}
\end{table*}

\begin{figure}[t]
\vspace{-0.3cm}
\centering
\subfloat[Edges Embedding Output]{   
\hspace{-0.3cm}\includegraphics[height=2.7cm]{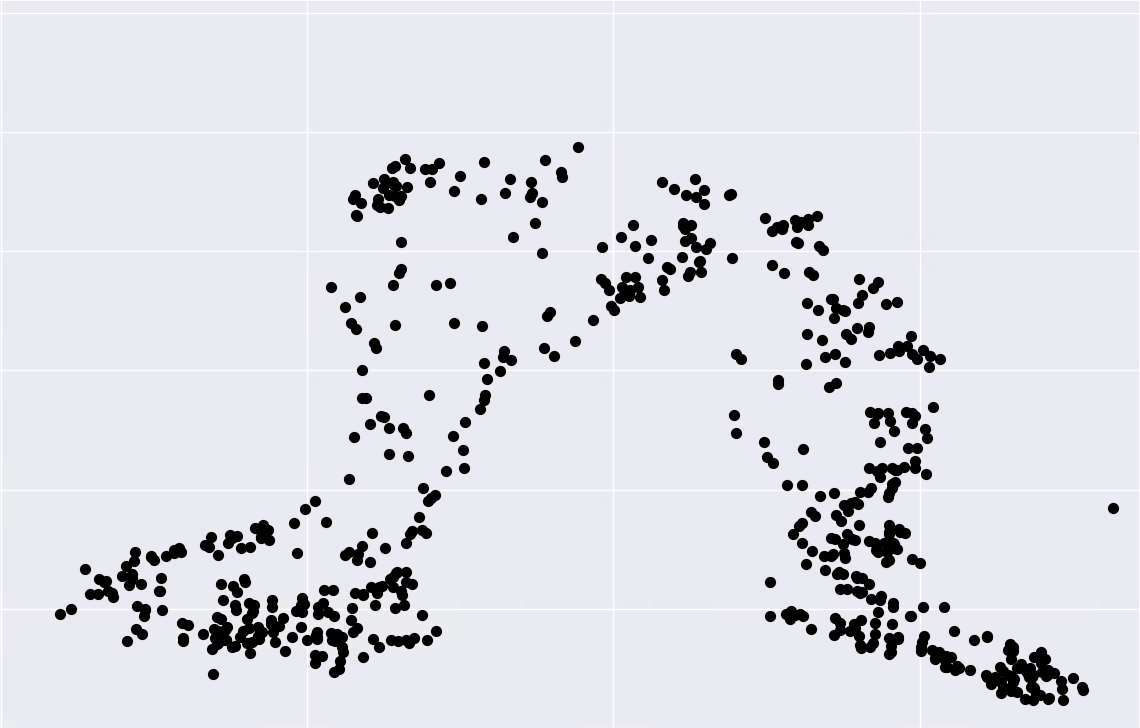}
}
\subfloat[Edges Clustering  Output]{  \includegraphics[height=2.7cm]{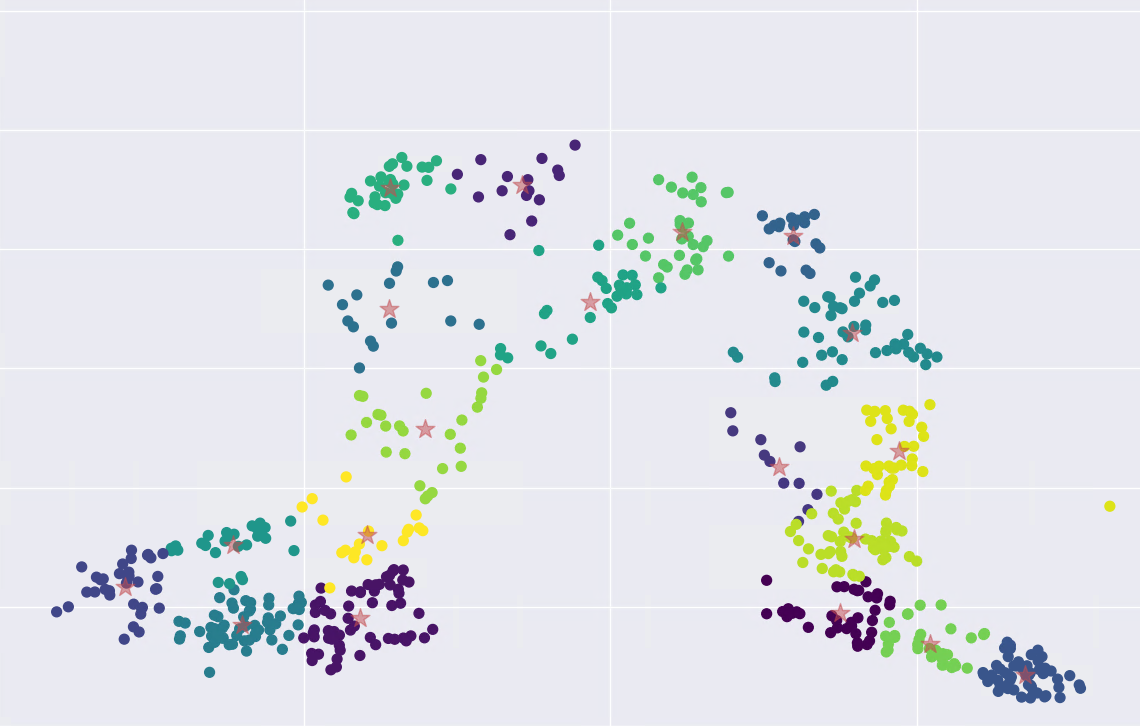}}
  \\
  \centering
    \subfloat[Edges and Attributes Clustering  Output]{  \hspace{-0.2cm}\includegraphics[width=4.2cm]{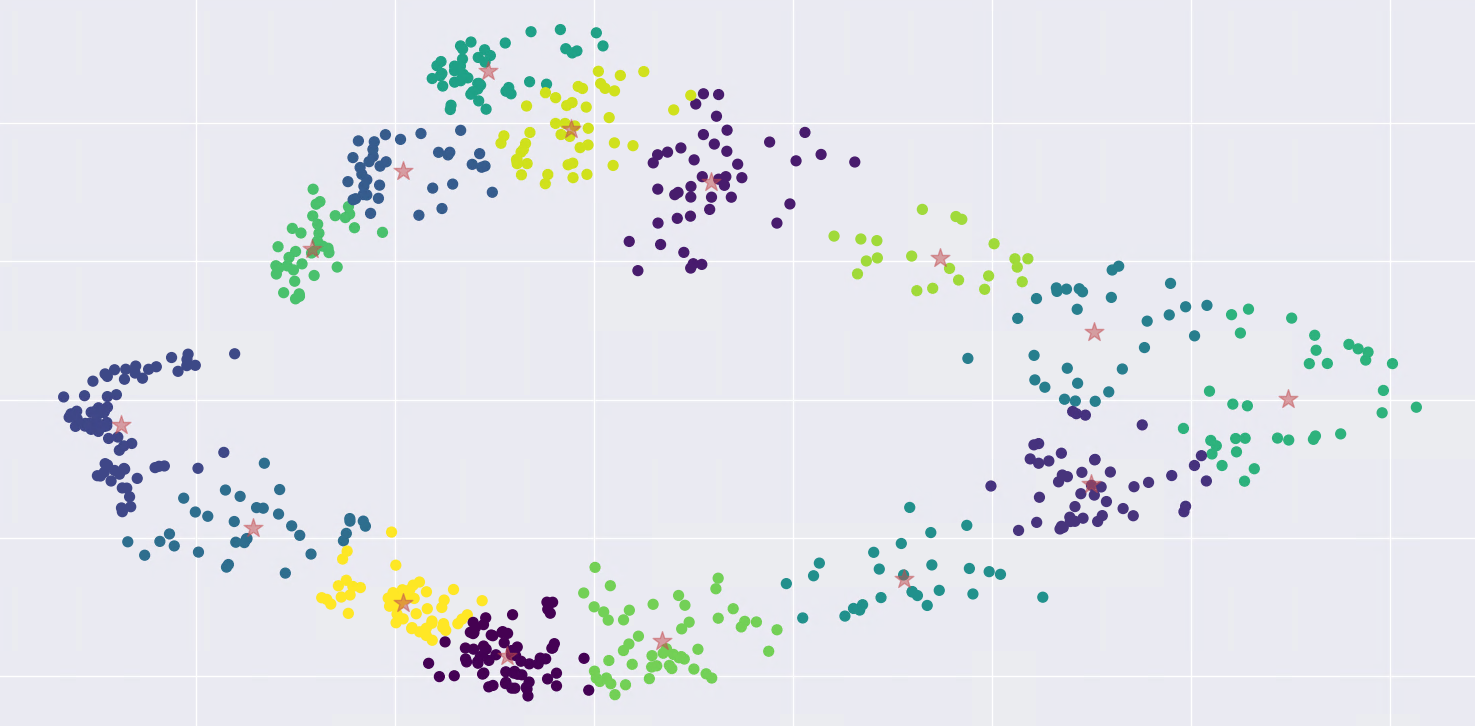}}\hspace{0.2cm}
    \subfloat[Attributes Clustering  Output]{  \hspace{-0.3cm}\includegraphics[width=4.2cm]{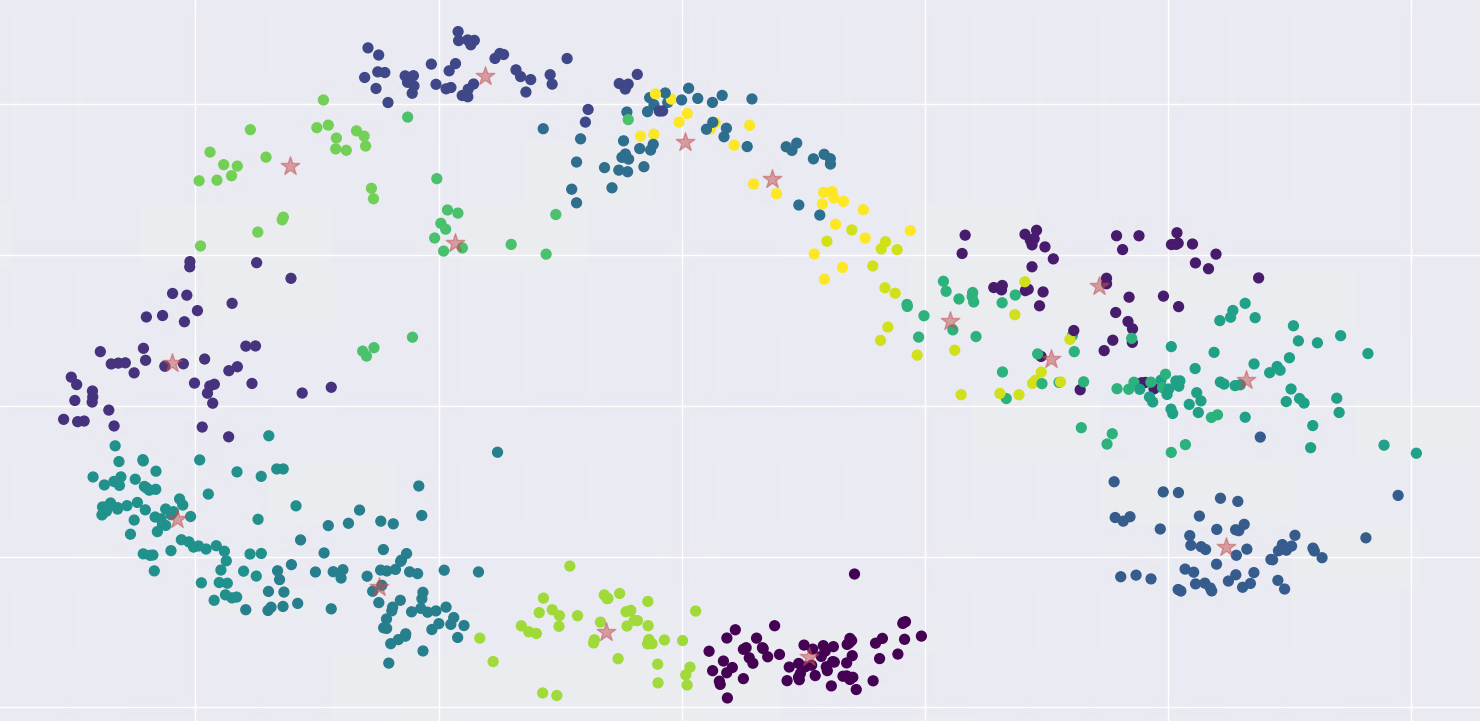}}
    \caption{Embedding output and clustering output for different types of embeddings.}  \vspace{-0.3cm}
  \label{2figures} 
\end{figure}

\subsection{Simulation Results}
For the embedding phase, we employ Co-Embedding Attributed Networks (CAN) algorithm by Meng. et al.\cite{Meng2019CoEmbeddingAN}. As CAN is a state-of-the-art model that learns low-dimensional representations of both attributes and nodes in the same semantic space, we propose to utilize it in order to embed the three approaches: 1) SFOR relationships only, 2) SIoT device's extracted attributes, and 3) both SFOR relationships and device's attributes at the same time. For the first approach, we neutralize the effect of the attributes in the embedding by setting them all to same value for all of the devices, and consequently removing their effect in the embedding. Similarly, in the second approach, we neutralize the effect of the SIoT graph by transforming the graph to a unweighted fully connected graph. In order to choose the number of feature embedding dimensions, we perform a preliminary simulation where we vary the number of dimensions and compute the embedding accuracy. The result of this simulation is illustrated in Fig.~\ref{2other}-(a). As we can notice, the accuracy of the embedding is at its highest when the values of the number of embedding dimension is 32 for all of the approaches. The result of the embedding for up to 250 epochs is also illustrated in Fig.~\ref{2other}-(b). As we can see, after 250 epochs, the SFOR achieved lower than 95\% accuracy while the other two embedding techniques have more than 95\% accuracy. 


After the embedding procedure, we proceed with the clustering phase. We utilize k-means, which is a clustering method based on vector quantization to partition the IoT devices into clusters, in which each device belongs to the cluster with the nearest mean in euclidean distance. Fig.~\ref{2figures}-(b), Fig.~\ref{2figures}-(c), and Fig.~\ref{2figures}-(d), show the clustering results in the embedding space. Each point represents an IoT device, and the color reflects the group to which that device belongs in terms of similarity. The distance between two nodes represents the distance between the IoT devices. Moreover, the more two nodes are similar and close in the social network, the closer they are in the embedding space. The cluster centers, also known as clusters centroid, are represented as red stars. The color of nodes represents their assigned cluster. The nodes having similar colors are assigned to the same community. After achieving the clustering, we proceed with the service lookup procedure as mentioned in Section~\ref{sec3}. During this phase, the service discovery approach selects, from the pool of clusters, candidate IoT devices to complete the required service.  The overall performances of the three approaches are included in Table.~\ref{tablescaract}. In fact, we compute the minimum number of embedding epochs needed to achieve 95\% accuracy, the accuracy of the embedding algorithms at 500 epochs, the average number of candidate devices from the chosen cluster, the similarity degree to the service requester's SIoT relations, and the similarity degree to the server requester's required characteristics. We notice that the attribute embedding approach converged faster than the other two approaches. The edge-attributes embedding took on average 730 epochs to converge. The highest accuracy at 500 epochs is obtained by the attributes embedding approach while the lowest accuracy is achieved by the edge-attributes approach. The candidate devices chosen by the attribute embedding approach are the most close devices to the requester's required characteristics. The edge-attributes embedding achieves a trade-off between SIoT relations and required characteristics by achieving close characteristics to the required ones with strong relationships to the service requester.


\vspace*{-0.15cm}

\section{Conclusion}\label{sec5}
In this paper, we have addressed the issue of service discovery in large-scale social IoT by proposing a resource allocation neural model based on GNNs. We have utilized the social relationships formed between the devices to reduce the search space of any entity looking to acquire a service from its peers. The proposed low-complexity resource allocation approach surpasses standardization issues and embeds characteristics of the SIoT graph for eventual clustering process. Simulation results applied on a real-world dataset show its significant efficiency in downgrading the complexity of large-scale IoT networks for better structure understanding and manipulation.

\balance
\bibliographystyle{IEEEtran}
\bibliography{references}
\end{document}